%% file: egpaper_final.tex
\documentclass[10pt,onecolumn,letterpaper]{article}
\usepackage{cvpr}
\usepackage{times}
\usepackage{epsfig}
\usepackage{graphicx}
\usepackage{amsthm}
\usepackage{amsmath}
\usepackage{amssymb}
\usepackage[breaklinks=true,bookmarks=false]{hyperref}

\numberwithin{equation}{subsection}
\usepackage{datetime}

\usepackage[utf8]{inputenc} 
\usepackage[T1]{fontenc}    
\usepackage{hyperref}       
\usepackage{url}            
\usepackage{booktabs}       
\usepackage{amsfonts}       
\usepackage{nicefrac} 

\usepackage{microtype}      
\usepackage{xcolor}         
\usepackage{booktabs}
\usepackage{amssymb}
\usepackage{mathtools}
\usepackage{amsthm}
\usepackage{xcolor}
\usepackage{multirow} 
\usepackage{subcaption} 
\usepackage[a4paper,margin=1in,headheight=15pt,headsep=24pt]{geometry}

\usepackage{fancyhdr}

\cvprfinalcopy 
\def\cvprPaperID{****} 
\allowdisplaybreaks

\begin{document}
\lhead{}

\title{LARGO: Low-Rank Regulated Gradient Projection for Robust Parameter Efficient Fine-Tuning}
\author{
\textbf{Haotian Zhang}$^{1,2}$\quad
\textbf{Liu Liu}$^{1,2*}$\quad
\textbf{Baosheng Yu}$^{3}$\quad
\textbf{Jiayan Qiu}$^{4}$\\
\textbf{Yanwei Ren}$^{1,2}$\quad
\textbf{Xianglong Liu}$^{1}$\\
[0.5em]
$^1$School of Artificial Intelligence, Beihang University\\
$^2$Hangzhou International Innovation Institute, Beihang University\\
$^3$Nanyang Technological University \quad
$^4$University of Leicester 
}
\maketitle

\begingroup
\renewcommand\thefootnote{*}
\footnotetext{Corresponding author: \texttt{liuliubh@buaa.edu.cn}}
\endgroup

\input{NIPStext/0abs}
\input{NIPStext/1Intr}
\input{NIPStext/2Rel}
\input{NIPStext/3Prel}

\input{NIPStext/4V2Method}
\input{NIPStext/5Exp}

\input{NIPStext/7Con}
{
\bibliographystyle{ieee}
\bibliography{egbib}
}
\newpage
\input{NIPStext/6sup}

\end{document}

%% file: NIPStext/0abs.tex
\begin{abstract}
The advent of parameter-efficient fine-tuning  methods has significantly reduced the computational burden of adapting large-scale pretrained models to diverse downstream tasks. However, existing approaches often struggle to achieve robust performance under domain shifts while maintaining computational efficiency. To address this challenge, we propose Low-rAnk Regulated Gradient Projection (LARGO) algorithm that integrates dynamic constraints into low-rank adaptation methods. Specifically, LARGO incorporates parallel trainable gradient projections to dynamically regulate layer-wise updates, retaining the Out-Of-Distribution robustness of pretrained model while preserving inter-layer independence. Additionally, it ensures computational efficiency by mitigating the influence of gradient dependencies across layers during weight updates. Besides, through leveraging singular value decomposition of pretrained weights for structured initialization, we incorporate an SVD-based initialization strategy that minimizing deviation from pretrained knowledge. Through extensive experiments on diverse benchmarks, LARGO achieves state-of-the-art performance across in-domain and out-of-distribution scenarios, demonstrating improved robustness under domain shifts with significantly lower computational overhead compared to existing PEFT methods. 
The source code will be released soon.
\end{abstract}

%% file: NIPStext/1Intr.tex
\section{Introduction}
Recent advancements in large-scale pretrained vision models \cite{dosovitskiy2021an, chen2021crossvit,liu2022swin,el-nouby2024scalable} have revolutionized the field of image classification. These models, trained on extensive datasets, encode highly transferable representations and achieve state-of-the-art performance across diverse image classification benchmarks \cite{deng2009imagenet, russakovsky2015imagenet}. Despite these successes, adapting pretrained models to specific downstream tasks poses unique challenges, particularly in maintaining robust performance under domain shifts. \cite{ganin2016domain,saito2018open,tzeng2017adversarial}.


As large models become standard, parameter-efficient fine-tuning (PEFT) has emerged as the preferred approach, adapting only a small subset of parameters through techniques like Adapters~\cite{lu2023uniadapter}, Prompt Tuning~\cite{lester2021power}, BitFit~\cite{zaken2021bitfit}, and LoRA~\cite{hu2021lora,liu2024dora,balazy2024lora,biderman2024lora}. However, existing robustness methods are not designed for PEFT, limiting their applicability and efficiency in this setting~\cite{wortsman2022robust,huang2025directional}. This motivates the need for scalable, PEFT-compatible strategies for robust adaptation under distribution shift.

To address the challenge of out-of-distribution (OOD) robustness, methods using different regularization, such as  Maximum Absolute Row Sum (MARS)  \cite{gouk2021distancebased, xuhong2018explicit} and Trainable Projection Gradient Method (TPGM) \cite{tian2023trainable}  have been developed, employing hard constraints through projection-based approaches to enforce robustness during fine-tuning.In particular, MARS ensures that the fine-tuned model remains within a predefined distance from the pretrained weights, effectively controlling the trade-off between task adaptation and model stability. This approach guarantees that OOD robustness is not sacrificed for task-specific performance, but it often comes with increased computational cost due to the need for explicit constraint enforcement. 
TPGM introduces bi-level optimization to dynamically learn layer-wise constraints,and enables the model to adapt to domain-specific challenges while preserving pretrained knowledge, offering flexibility and efficiency in OOD settings. 
Additionally, they may introduce unnecessary computational overhead and lack the flexibility needed for seamless integration with large-scale models.

To address these gaps, motivated by the need for fine-tuning methods that are both computationally efficient and capable of enhancing OOD robustness within the PEFT framework,
We proposed \textbf{LARGO}, Low-rAnk Regulated Gradient Projection algorithm that enhances the PEFT framework by integrating trainable projection radii into the LoRA framework, enabling the independent learning of layer-wise constraints for each PEFT matrix. This mechanism ensures that weight updates are not influenced by gradient dependencies across layers, significantly improving robustness under domain shifts. Additionally, we introduced an \textbf{SVD-based initialization method}, which leverages singular value decomposition of pretrained weights to initialize the trainable parameters. This approach ensures minimal deviation from the pretrained model, enhancing computational efficiency while preserving the integrity of pretrained representations.
Furthermore, to efficiently adapt to downstream tasks, we introduce an SVD-based initialization method within the LARGO algorithm. Unlike Kaiming initialization \cite{he2015delving} that using Gaussian vector, this method decomposes the original weight matrix and scales the singular values of the decomposed matrices to a smaller magnitude.

The contributions of this paper are as follows:
\begin{itemize}
    \item We propose  a new algorithm that incorporates trainable projection radii into the LoRA framework. This method independently learns layer-wise constraints for each PEFT matrix, ensuring that the regulation of weight updates is not influenced by gradient dependencies across layers, improving robustness under domain shifts. 
    \item We develop an SVD-based initialization method, which leverages singular value decomposition of pretrained weights to initialze trainable parameters in a structured and computationally efficient manner, ensuring minimal deviation from pretrained weights.
    \item Comprehensive experiments on DomainNet and ImageNet demonstrate that the proposed LARGO algorithm achieve superior OOD performance while maintaining competitive in-distribution accuracy, with minimal hyperparameter tuning.
\end{itemize}

%% file: NIPStext/2Rel.tex
\section{Related Works}  \label{sec_relatedworks}

\textbf{Parameter-Efficient Fine-Tuning} Parameter-efficient fine-tuning (PEFT) has become a core strategy for adapting large models under resource constraints. Delta-tuning frameworks provide a unified categorization of PEFT approaches \cite{ding2023parameter}, while methods like Aurora leverage prompt-based transfer to reduce complexity in low-resource settings \cite{wang2023parameter}. Adapter-based techniques such as UniAdapter integrate unimodal and multimodal modules for efficient adaptation \cite{lu2023uniadapter}. Among PEFT methods, Low-Rank Adaptation (LoRA) and its extensions—DoRA and PiSSA—remain prominent for injecting low-rank updates into frozen pre-trained weights \cite{hu2021lora, liu2024dora, meng2024pissa}. These techniques improve efficiency while preserving performance.
Consistency-based methods such as PACE and the approach by Roy et al. enhance generalization through perturbation-invariant training \cite{ni2024pace, roy2023consistency}. LISA further reduces memory usage by freezing middle layers via layerwise importance sampling \cite{pan2024lisa}. While these methods successfully reduce parameter overhead, they often overlook robustness under distribution shift. Consequently, new strategies are needed that maintain both efficiency and adaptability in OOD settings.

\textbf{Adversarial Robustness} Robust fine-tuning is critical for maintaining generalization under distribution shifts and adversarial perturbations. Surgical fine-tuning improves OOD performance by selectively updating specific layers \cite{lee2023surgical}, and Kumar et al. show that full fine-tuning may underperform linear probing when pre-trained features are already strong \cite{kumar2022finetuning}. Projection-based methods such as MARS and FTP control weight drift to improve generalization \cite{gouk2021distancebased, tian2024fast}. TPGM extends this idea with layer-wise constraints for improved robustness \cite{tian2023trainable}.
In multimodal settings, CaRot combines constrained contrastive learning and EMA-based self-distillation to improve both accuracy and calibration \cite{jang2024model}. Large-scale empirical studies on PEFT \cite{lialin2023scaling} reveal that many methods face scalability bottlenecks, particularly under OOD shifts. Although recent advances offer promising results, most approaches remain computationally costly or brittle in complex domains, underscoring the need for robust and scalable PEFT strategies.

%% file: NIPStext/3Prel.tex
\section{Preliminary}\label{sec_prelimnary}
\textbf{Notation}: For  a matrix $\mathbf{X}$,  we use $\|\textbf{X}\|_2$ and $\|\textbf{X}\|_1$ to denote the $L_2$ and $L_1$  norm, respectively. 
We use $\partial$ and $\nabla$ to denote its partial gradient and gradient. 

\subsection{Low-Rank Adaptation}
Parameter-based tuning adapts pretrained models by modifying parameters, such as weights or biases, in a structured manner. This typically involves scaling the feature map by a weight matrix and adding a bias term for refinement. 
A representative example is Low-Rank Adaptation (LoRA) \cite{hu2021lora}, which decomposes weight updates, $\Delta \textbf{W}$, into low-rank matrices. Instead of updating the full $\Delta \textbf{W}$, LoRA learns two smaller matrices through backpropagation, capturing the updates efficiently.
To understand LoRA, consider the typical weight update during fine-tuning:Let $\mathbf{W}$ be the weight matrix. The update $\Delta \textbf{W}$ is computed via backpropagation as:
\begin{align} 
    \Delta \textbf{W} = \eta (-\nabla \mathcal{L}_W),
\end{align} 
where $\nabla \mathcal{L}_W$ represents the gradient of the loss function with respect to $\textbf{W}$ and $\eta$ is the learning rate. 


LoRA's key insight lies in the concept of low intrinsic dimensionality. This refers to the observation that many tasks can be effectively represented in a lower-dimensional space without losing significant information. Based on this idea, LoRA decomposes the weight update $\Delta \textbf{W} \in \mathbb{R}^{\text{A} \times \text{B}}$ into two smaller matrices: 
\begin{align} 
    \Delta \textbf{W} = \textbf{W}_A \textbf{W}_B,
\end{align} 
where $\textbf{W}_A$ is an $A \times r$ matrix and $\textbf{W}_B$ is an $r \times B$ matrix, with $r \ll \min\{A, B\}$. Importantly, the original weight matrix $\textbf{W}$ remains fixed, while $\textbf{W}_A$ and $\textbf{W}_B$ are optimized. This decomposition reduces the number of trainable parameters significantly while maintaining the model's ability to adapt effectively. A schematic representation of this method is provided in the figure below.

\subsection{Distance-Based Regularization and Trainable Projected Gradient}
\textbf{Distance-Based Regularization}:
To achieve robust generalization performance, the MARS method emphasizes that the weights of fine-tuned models should remain close to their pretrained counterparts. Specifically, MARS defines a fixed distance constraint, ensuring that the learning process is restricted to a weight space within this predefined distance from the pretrained weights. To enforce this, MARS employs a two-step strategy: 1)  Using projected sub-gradient methods to impose a hard constraint; 2) Augmenting the standard cross-entropy objective with an additional penalty term that quantifies the distance between the pretrained and fine-tuned weights.
This approach guarantees that the constraints are upheld, even when heuristic optimization methods are applied to train the network.
A key component of MARS is the projection function, which ensures that updated weights do not violate the predefined constraints. For classic subgradient descent, convergence to a stationary point requires the projection function to perform a Euclidean projection. Formally, this can be expressed as:
\begin{align}
    \pi(\widehat{\textbf{W}})=\underset{\textbf{W}}{\arg \min } \frac{1}{2}\|\textbf{W}-\widehat{\textbf{W}}\|_2^2 \quad
    \text { s.t. } g(\textbf{W}) \leq 0,
\end{align}
where $\widehat{\textbf{W}}$ represents the updated weights that may violate the constraints, $\textbf{W}$ represents the projected weights, and $g(\cdot)$ is a convex function defining the constraints.


Furthermore, MARS translates the Frobenius distance constraint into a collection of $l_1$ distance constraints applied row-wise to the weight matrix. For a given row of weights(for example $n$), the projection function can be expressed as: 

When the $l_1$ distance constraints is employed, the constraint function $g(\textbf{w}_i), i\in [n]$ is defined as
\begin{align}
    g(\textbf{w}_i)=\|\textbf{w}_i-\textbf{w}_i^{(0)}\|_1-\gamma \leq 0,
\end{align}
where $\textbf{w}_i^{(0)}$refers to the initial value of ith row of matrix the optimal solution becomes:
\begin{align} 
    \pi_1(\widehat{\textbf{w}}_i-{\textbf{w}}_i^{(0)}, \gamma)\!=\!\frac{1}{\text{max} (1, {\|\widehat{{\textbf{w}}}_i\!-\!{\textbf{w}}_i^{(0)}\|_1}/{\gamma})} (\widehat{{\textbf{w}}}_i\!-\!{\textbf{w}}_i^{(0)}). 
\end{align}
The overall projection function for the entire weight matrix is then obtained by applying $\pi_1$ to each row individually and translating the results back into the weight space:  
\begin{align}\label{Eq:projection1}
    \pi_{\infty}\left(\textbf{W}^{(0)}, \widehat{\textbf{W}}, \gamma\right)\!=\!\left[\begin{array}{c}
    \pi_1(\widehat{{\textbf{w}}}_1-{\textbf{w}}_1^{(0)}, \gamma)+{\textbf{w}}_1^{(0)} \\
    \vdots \\
    \pi_1(\widehat{{\textbf{w}}}_n-{\textbf{w}}_n^{(0)}, \gamma)+{\textbf{w}}_n^{(0)}
    \end{array}\right].
\end{align}

While projection-based methods like those used in MARS provide strong theoretical guarantees for satisfying constraints, they may introduce implementation challenges, especially in scenarios requiring efficient optimization. In practice, penalty-based approaches are more commonly adopted due to their compatibility with automatic differentiation frameworks. However, penalty-based methods often lack the rigorous assurances of constraint fulfillment that projection-based approaches inherently provide.

\textbf{Trainable Projected Gradient Method}:
TPGM framework introduces a novel mechanism to enforce robust fine-tuning by maintaining projection radii and distance constraints for each layer, ensuring that the fine-tuned model remains close to the pretrained model in weight space. These constraints are dynamically enforced through weight projections, as demonstrated in the algorithms. To determine the optimal set of projection radii, TPGM employs a bi-level optimization strategy, enabling the automatic and end-to-end learning of layer-specific constraints. Theoretically, this bi-level formulation is critical to achieving tailored constraints for each layer, providing a strong foundation for the method. Empirically, TPGM demonstrates superior out-of-distribution (OOD) performance compared to existing fine-tuning methods, while maintaining competitive in-distribution (ID) accuracy—all with minimal hyper parameter tuning effort. 
\begin{align}
    \widehat{\textbf{W}}_{t+1} =& \textbf{W}_{t}- \eta_t \nabla_W \mathcal{L}(x, y; \textbf{W}_{t}), \quad x, y \in \mathcal{D}_{\text{tr}}\\
    \widetilde{\textbf{W}}_{t+1}=&\Pi(\textbf{W}_0, \widehat{\textbf{W}}_{t+1}, \gamma_{\tau}),
    \\
    \gamma_{\tau+1} =& \gamma_\tau - \alpha \nabla_\gamma \mathcal{L}(x, y; \widetilde{\textbf{W}}_{t+1}), \quad x, y \in \mathcal{D}_{\text{val}}\
    \\
    \textbf{W}_{t+1} =& \Pi(\textbf{W}_0, \widehat{\textbf{W}}_{t+1}, \gamma_{t+1}),
\end{align}
where $\Pi(\cdot)$ is defined in Eq. \eqref{Eq:projection1}, 



A major drawback of TPGM is its computational cost, especially with large-scale models. The bi-level optimization requires frequent updates to the projection radii, increasing training time and resource usage. TPGM's reliance on projections also limits its scalability in parameter-efficient fine-tuning (PEFT) settings, where efficiency and robustness are crucial.These limitations highlight the need for fine-tuning methods that balance computational efficiency with robust OOD generalization,which motivate the development of our approach: by  integrating dynamic constraints while minimizing computational overhead, we provide a more efficient and scalable solution within the PEFT framework.

%% file: NIPStext/4V2Method.tex
\begin{figure*}[t]
	\centering	\includegraphics[width=0.85\linewidth]{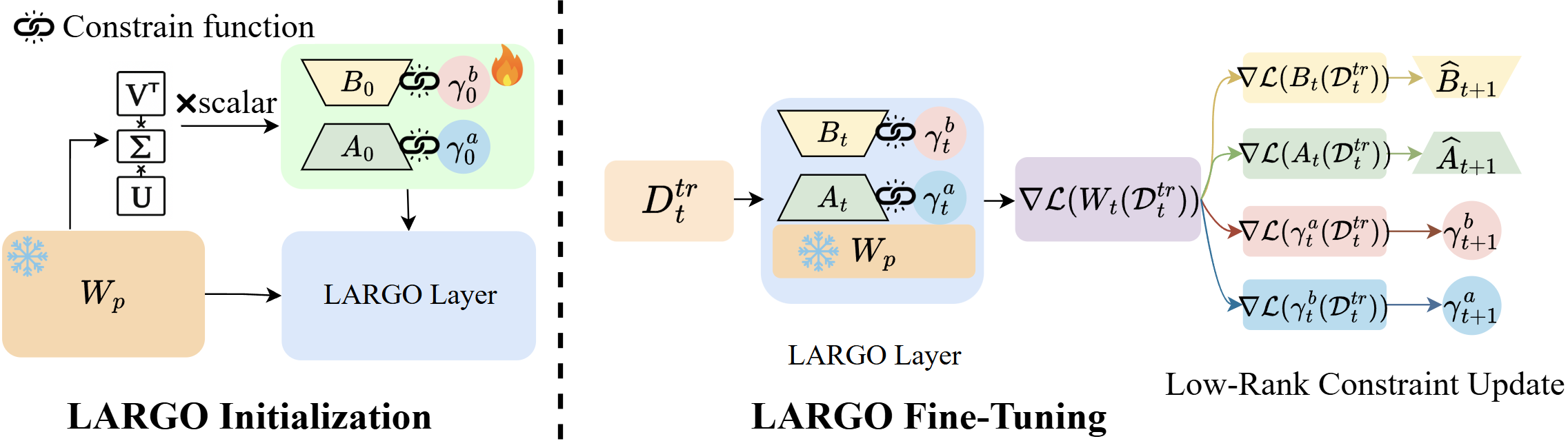}
    \caption{Overview of the \textbf{LARGO} method. The process computes the gradient for whole LARGOlayers, followed by parallel updates of the low-rank matrices $\mathbf{A}$ and $\mathbf{B}$ and their respective trainable scalars $\gamma^a$ and $\gamma^b$. This  constraint update mechanism dynamically modulates weight adjustments, ensuring efficient and balanced adaptation while preserving pretrained robustness. }
    \label{Figure:LARGO_figure}
\end{figure*}

\section{Proposed Method} \label{sec_proposedmethod}
\subsection{Low-Rank Regulated Gradient Projection}
To enhance the robustness of downstream models while retaining the advantages of Parameter-Efficient Fine-Tuning methods, we propose a novel approach, Low-Rank Regulated Gradient Projection, \textbf{LARGO}, which focuses on assimilating the robustness of pretrained models by introducing learnable weight constraints into certain PEFT techniques, ensuring effective fine-tuning without compromising the model's inherent stability.

\subsubsection{The update of parameters}
During the update process at epoch $t$, the partial gradients of the loss function $\mathcal{L}$ are computed as follows:
\begin{align}
    \frac{\partial \mathcal{L}}{\partial {\textbf{W}_{t-1}}}=\frac{\partial \mathcal{L}}{\partial\left({\textbf{W}}_0+{\textbf{A}_{t-1} \textbf{B}_{t-1}}\right)}, 
    \quad
    \frac{\partial \mathcal{L}}{\partial {\textbf{A}_{t-1}}}=\frac{\partial \mathcal{L}}{\partial {\textbf{W}_{t-1}}}  \textbf{B}_{t-1}^{T}, 
    \quad
    \frac{\partial \mathcal{L}}{\partial {\textbf{B}_{t-1}}}=& \textbf{A}_{t-1}^{T}  \frac{\partial \mathcal{L}}{\partial {\textbf{W}_{t-1}}}.
\end{align}
Based on the partial gradient, the updates become
\begin{align}
    \widehat{\textbf{A}}_t =\textbf{A}_{t-1} -\eta_{t-1} \frac{\partial \mathcal{L}}{\partial \textbf{A}_{t-1}}, 
    \quad
    \widehat{\textbf{B}}_t = \textbf{B}_{t-1} -\eta_{t-1} \frac{\partial \mathcal{L}}{\partial \textbf{B}_{t-1}}, 
    \quad
    \widehat{\textbf{W}}_t = \textbf{W}_0+\widehat{\textbf{A}}_t \widehat{\textbf{B}}_t .
\end{align}

During the training process of LoRA, we constrain the updates of the low-rank matrices $\textbf{A}$ and $\textbf{B}$ by introducing trainable scalars $\gamma^a$ and $\gamma^b$ that are updated during training, which dynamically modulate the weight updates for each matrix. 
\begin{align}
    \|\textbf{a}_i-\textbf{a}_i^{(0)}\|_1-\gamma_a \leq 0,
    \|\textbf{b}_i-\textbf{b}_i^{(0)}\|_1-\gamma_b \leq 0,
\end{align}
where $\textbf{a}_i^{(0)}$ and $\textbf{b}_i^{(0)}$ are the rows of $\textbf{A}_0$ and $\textbf{B}_0$.

By doing so, we can control the magnitude of the updates applied to the weight matrices, providing a more flexible and adaptive mechanism that balances between the preservation of pretrained knowledge and the need for task-specific adaptation. This dynamic control allows for greater control over the fine-tuning process, which is especially useful in scenarios where computational efficiency and robustness are equally critical.

The trainable scalars $\gamma^a$ and $\gamma^b$ independently regulate the weight updates of matrices $\textbf{A}$ and $\textbf{B}$ for each layer, allowing the constraints to be learned and adjusted without being influenced by gradient dependencies across layers. This independence ensures that each PEFT matrix’s updates are modulated according to its specific requirements, rather than being constrained by inter-layer relationships. 
As shown below, the process of weight update is modified by the scalars:
\begin{align}
    \textbf{W}_t=(\widehat{\textbf{W}}_t-\textbf{W}_0) \frac{\gamma^a_t\gamma^b_t}{\|\widehat{\textbf{W}}_t-\textbf{W}_0\|_1} + \textbf{W}_0,
\end{align}
since LoRA has decomposed the $\widehat{\textbf{W}}_t-\textbf{W}_0$ into $\widehat{\textbf{A}}_t \widehat{\textbf{B}}_t$, equivalently:
\begin{align}
    \textbf{W}_t = (\widehat{\textbf{A}}_t \widehat{\textbf{B}}_t) \frac{\gamma^a_t \gamma^b_t}{\|\widehat{\textbf{A}}_t\|_1\|\widehat{\textbf{B}}_t\|_1} + \textbf{W}_0.
    \label{eq:largo_ab}
\end{align}
\subsubsection{The update of parameter $\gamma$}
In Eq. \eqref{eq:largo_ab}, $\textbf{W}_t$ becomes a function of $\widehat{\textbf{A}}_t$, $\widehat{\textbf{B}}_t$, and $\gamma_t$, which allows the gradients to propagate back to $\gamma^a_t$ and $\gamma^b_t$. The gradient flow through the scalars enables us to adjust the learning rate of the weight updates dynamically, which is key for handling the trade-offs between efficiency and robustness. By the chain rule, the gradients with respect to $\gamma^a_t$ and $\gamma^b_t$ are derived as:
\begin{align}
    \frac{\partial \mathcal{L}}{\partial \gamma^a_t} = \frac{\partial \mathcal{L}}{\partial \textbf{A}_t} \frac{\partial \textbf{A}_t}{\partial \gamma^a_t} =t \frac{\partial \mathcal{L}}{\partial \textbf{W}_t}  {\textbf{B}_t^{\top}} \frac{\widehat{\textbf{A}}_t}{\|\widehat{\textbf{A}}_t\|_1}, 
    \quad
    \frac{\partial \mathcal{L}}{\partial \gamma^b_t} = \frac{\partial \mathcal{L}}{\partial \textbf{B}_t} \frac{\partial \textbf{B}_t}{\partial \gamma^b_t} = t{\textbf{A}_t^{\top}}  \frac{\partial \mathcal{L}}{\partial \textbf{W}_t} \frac{\widehat{\textbf{B}}_t}{\|\widehat{\textbf{B}}_t\|_1}.
\end{align}

With these gradients, the scalar parameters are updated as follows:
\begin{align}
\gamma^a_{t+1} = \gamma^a_t - \frac{\partial \mathcal{L}}{\partial \gamma^a_t} = \gamma^a_t - t\frac{\partial \mathcal{L}}{\partial \textbf{W}_t}  {\textbf{B}_t^{\top}} \frac{\widehat{\textbf{A}}_t}{\|\widehat{\textbf{A}}_t\|_1}, 
\label{gamma_a_update}
\\
\gamma^b_{t+1} = \gamma^b_t - \frac{\partial \mathcal{L}}{\partial \gamma^b_t} = \gamma^b_t - t{\textbf{A}_t^{\top}}  \frac{\partial \mathcal{L}}{\partial \textbf{W}_t} \frac{\widehat{\textbf{B}}_t}{\|\widehat{\textbf{B}}_t\|_1}. 
\label{gamma_b_update}
\end{align}

This approach contrasts with methods that enforce global dependencies between layers, where updates across layers are interconnected, often leading to increased complexity and computational cost. By independently learning layer-wise constraints, our method provides a more flexible and efficient solution, allowing for better scalability and reduced overhead. The scalars $\gamma^a_t$ and $\gamma^b_t$ thus dynamically control the strength of the adaptation, balancing the need for task-specific exploration while preserving the pretrained model’s knowledge, with significantly lower computational expense.
Finally, the weights for the next epoch are updated as:
\begin{align}
\textbf{W}_{t+1} = (\widehat{\textbf{A}}_{t+1} \widehat{\textbf{B}}_{t+1}) \frac{\gamma^a_{t+1} \gamma^b_{t+1}}{\|\widehat{\textbf{A}}_{t+1}\|_1\|\widehat{\textbf{B}}_{t+1}\|_1} + \textbf{W}_0.
\end{align}

The overall process is illustrated in Figure~\ref{Figure:LARGO_figure}. By combining learnable scalar constraints with LoRA, this approach preserves the robustness of pretrained models while maintaining the efficiency of PEFT methods. The trainable scalars $\gamma^a$ and $\gamma^b$ provide a dynamic mechanism for adapting the strength of updates, allowing the model to effectively adjust to both in-domain performance and out-of-distribution generalization. This method addresses the need for both task-specific adaptation and computational efficiency, offering a scalable solution for fine-tuning large models. Moreover, the adaptability of the scalars ensures that the updates are tailored to the task at hand, without sacrificing the model's ability to generalize to unseen distributions. As a result, our approach not only improves OOD robustness but also maintains the high efficiency characteristic of PEFT methods, making it a promising solution for fine-tuning pretrained models in challenging scenarios.

\subsection{SVD Initiazation}
To better preserve the out-of-distribution (OOD) robustness of the original model, we propose a novel initialization method leveraging Singular Value Decomposition (SVD). This method decomposes the original weight matrix and scales the singular values of the decomposed matrices to a small magnitude, using these scaled matrices as the initialization for $\textbf{A}_0$ and $\textbf{B}_0$ in LoRA. By utilizing this approach, we ensure that the pretrained model’s inherent robustness is maintained, while also enabling efficient adaptation to downstream tasks. 
Specifically, for each layer using SVD initialization, we apply SVD to its weight matrix $\textbf{W}_0 \in \mathbb{R}^{n \times d}$, decomposing it into three components: $\textbf{\textbf{U}} \in \mathbb{R}^{n \times n}$, $\textbf{S} \in \mathbb{R}^{n \times d}$, and $\textbf{V}^T \in \mathbb{R}^{d \times d}$. We then select the top $r$ singular values by clipping the matrices $\textbf{U}$, $\textbf{S}$, and $\textbf{V}^T$ to their leading $r$ components, resulting in $\textbf{U}_r \in \mathbb{R}^{n \times r}$, $\textbf{S}_r \in \mathbb{R}^{r \times r}$, and $\textbf{V}_r^T \in \mathbb{R}^{r \times d}$.
To ensure the initial values of $\textbf{A}_0$ and $\textbf{B}_0$ are close to zero, we introduce a manual scaling parameter $s$ and  based scaling parameter $s_r$ based on singular values of $\textbf{S}_r$ using the following formula:
\begin{equation}
s_r=\frac{r}{\|\textbf{S}_r\|_2^2}.
\end{equation}

This scaling ensures that the initialization remains close to the pretrained weights while being computationally efficient. The final initialization matrices for $\textbf{A}_0$ and $\textbf{B}_0$ are computed as:
\begin{align}
    \textbf{A}_0 = t\textbf{U}_r \sqrt{s_r}, \quad \textbf{B}_0 =t\sqrt{s_r}  \textbf{V}_r^T.
\label{FTP_init_formula}
\end{align}
Through scaling, we have
%
\begin{align}
    \textbf{A}_0 \leftarrow s t^2  \textbf{U}_r \frac{\sqrt{r}}{\|\textbf{S}_r\|_2}, \quad \textbf{B}_0 \leftarrow s t^2 \frac{\sqrt{r}}{\|\textbf{S}_r\|_2} \textbf{V}_r^T.
\label{FTP_init_formula}
\end{align}
By scaling the decomposition matrices in this way, we ensure that the model begins with an initialization that remains close to the pretrained weights, which allows for retaining OOD robustness from pretrained model and reserving potential for further task specific fine-tuning.

This initialization strategy ensures that the LoRA parameters $\textbf{A}_0$ and $\textbf{B}_0$ remain close to zero while maintaining the principal features of the original weight matrix. The scaled singular values allow the model to retain OOD robustness and avoid introducing large perturbations in the initial stage of training.


%% file: NIPStext/5Exp.tex
\section{Experiments}\label{sec_expriments}
\subsection{Datasets}
We utilized both DomainNet \cite{peng2019moment} (600K images across 345 classes) and ImageNet \cite{deng2009imagenet} as benchmark datasets to evaluate the performance of our method: 1)
\textbf{DomainNet} , which is a large-scale dataset spanning five distinct visual domains: \textit{Real}, \textit{Clipart}, \textit{Infograph}, \textit{Painting}, and \textit{Sketch}, enabling evaluation of both in-distribution (ID) and out-of-distribution (OOD) performance. We train on the \textit{Real} domain and use its validation split for ID evaluation. OOD generalization is assessed on the remaining four domains, which include stylized illustrations, schematic diagrams, artistic renderings, and abstract hand-drawn sketches.
%
%
2) \textbf{ImageNet} which is commonly used to assess both ID and OOD robustness. We fine-tune and validate our base model on \textit{ImageNet2012} dataset as ID dataset. The OOD evaluation  of ImageNet dataset series is conducted on \textit{ImageNetV2}~\cite{pmlr-v97-recht19a}, \textit{ImageNet-A}~\cite{hendrycks2021nae}, \textit{ImageNet-R}~\cite{Hendrycks_2021_ICCV}, and \textit{ImageNet-S}~\cite{wang2019learning}, which introduce shifts including distribution mismatch, adversarial instances, artistic renditions, and style perturbations that ensures sufficient distribution shift from different domains.

\subsection{Implementation}
We conduct fine-tuning on the ID subsets of the DomainNet and ImageNet families, and evaluate each model on corresponding ID and OOD benchmarks, with comparisons to a range of competitive baselines.In particular,
\noindent\textbf{DomainNet:} We use the released code from the previous work,FTP to fine-tune our baseline model.We apply LoRA to all attention blocks in our pre-trained ViT model and constrain all the trainable LoRA layers except for the last linear classification layers. 
For all experiments,we use SGD as the base optimizer with a weight decay of 5e-4. 
We trained our models for 150 epochs, respectively with a batch size of 256, and investigated the learning rate settings in FTP for each experiment. 
\noindent\textbf{ImageNet:} For Imagenet Datasets, We use the same pipeline as we used in DomainNet dataset, with the learning rate of 3e-5. We train all methods using SGD as base optimizer with a weight decay of 0.1, a cosine learning rate schedule, and a batch size of 256 for 30 epochs. Every experiment for both datasets was trained on 4 $\times$ NVIDIA V100s.
 
 \subsection{Experiment Results}

\input{NIPStext/table1}
\input{NIPStext/table2}
 
 We compared our method with several classic or SOTA approaches listed below:
     \textbf{Vanilla FT}: simply Fine-tune all parameters of the model.
     \textbf{Linear Probing}: Only modify the last linear layer of the model.
     \textbf{LP-FT \cite{osti_10472125}}: First performs Linear Probing for a fixed number of epochs to stabilize the model, followed by fine-tuning the entire model.
     \textbf{TPGM \cite{tian2023trainable}}:TPGM introduces a unique trainable constraint parameter for each layer, aiming to control the degree of parameter updates during fine-tuning.
     \textbf{FTP \cite{tian2024fast2}}: A variant of TPGM that reuses constraint parameters across layers to reduce redundancy and accelerate training. 

We evaluate LARGO on both the DomainNet and ImageNet datasets using the ViT-S architecture, comparing against strong baselines across in-distribution (ID) and out-of-distribution (OOD) settings. On \textbf{DomainNet} datasets, LARGO achieves an OOD average accuracy of 34.26\%, the highest among all methods, surpassing full fine-tuning (33.37\%), TPGM (32.76\%), and LoRA (30.31\%) while using only 0.42M trainable parameters. On the ID (real) split, LARGO reaches 73.92\%, slightly exceeding the full fine-tuning baseline of 73.68\%. On \textbf{ImageNet} datasets, LARGO achieves the best OOD average of 34.97\%, outperforming all baselines including full fine-tuning (33.70\%), LoRA (31.12\%), and TPGM (30.50\%). It sets new state-of-the-art results on ImageNet-R (35.07\%) and ImageNet-Sketch (33.84\%), while also achieving the highest ID accuracy of 76.66\%. Notably, these improvements are obtained with only 0.68M trainable parameters, representing a dramatic reduction in fine-tuning cost without sacrificing generalization. Overall, these results highlight LARGO's ability to achieve strong ID and OOD performance with minimal trainable parameters, offering a practical and scalable solution for efficient model adaptation. 
For further validation, we fine-tune and evaluate our method on the PaLI-Gemma-3B model using the VQAv2 dataset, comparing its performance against competitive baselines.
More results and details are available in 
Appendix A.


\subsection{Ablation study}
\noindent\textbf{Dataset split}
We split the DomainNet dataset into 10\%, 20\%, and 50\% splits and evaluated our method, LARGO, by fine-tuning a pre-trained model on each split separately. The performance of LARGO was compared with the current state-of-the-art (SoTA) methods under these three splits. Additionally, we analyzed the performance gap between SVD initialization and Kaiming initialization to assess their impact on model adaptation. We conduct  experiments on the following parameters:
    1) \textbf{\(\gamma\)}: The \(\gamma\) values in the tables represent the initial value of \(\gamma_a\) and \(\gamma_b\) used in Eq.\eqref{gamma_a_update} and \eqref{gamma_b_update}, which determine the expected constraint strength applied at each epoch.   
    2) $s$: Denotes the value of scalar defined in Eq. \eqref{FTP_init_formula}, which controls the scaling factor in the SVD initialization process.  
Furthermore, we conducted ablation experiments to investigate the effects of different \(\gamma\) values and the scaling weights of the SVD initialization on various datasets. The results of these experiments are detailed in the following sections.  

\noindent\textbf{Gamma Initialization}
To further investigate the impact of the initial value of \(\gamma\) on the updates in Eq. \eqref{gamma_a_update} and \eqref{gamma_b_update} across different datasets, we conducted a detailed analysis focusing on both Out-of-Distribution (OOD) and In-Distribution (real) average performance metrics. The results, presented in Table \ref{tab:gammaimpact_all}, reveal significant performance variations influenced by the initial choice of \(\gamma\).  
For the \textbf{10\% split}, a smaller initial \(\gamma\) value (\(1 \times 10^{-8}\)) achieves the highest OOD average (30.13) and a strong real average (66.55), demonstrating its effectiveness in enhancing OOD generalization with limited data. The trend holds for the \textbf{20\% split}, where \(\gamma = 1 \times 10^{-8}\) yields the best OOD average (32.48) and competitive real average (69.93). However, for the \textbf{50\% split}, the optimal \(\gamma\) shifts to \(1 \times 10^{-6}\) for the best OOD average (34.74), while \(\gamma = 1 \times 10^{-8}\) still delivers the highest real average (72.53). This shift suggests that larger datasets reduce the sensitivity to \(\gamma\), with slightly larger values better balancing OOD and in-domain performance.

These findings highlight several key insights:
1)
\textit{Smaller \(\gamma\) Values Enhance OOD Generalization:} Across all splits, smaller \(\gamma\) values (\(1 \times 10^{-8}\)) consistently yield higher OOD average scores, demonstrating their ability to improve domain generalization.  
2) \textit{Maintaining In-Domain Performance}: Despite enhancing OOD generalization, smaller \(\gamma\) values also maintain competitive in-domain performance, as evidenced by their strong real average scores.  
3) \textit{Dynamic Sensitivity to \(\gamma\)}: The optimal \(\gamma\) value varies slightly with dataset size, suggesting that the choice of \(\gamma\) should be tailored to the specific data split and task requirements.  
These results emphasize the importance of carefully selecting the initial \(\gamma\) value to balance OOD generalization and in-domain accuracy. Future work could explore \textbf{adaptive \(\gamma\) strategies} that dynamically adjust its value during training, enabling models to better navigate the trade-offs between OOD and in-domain performance across diverse datasets and scenarios.  




\begin{table}[hbpt]
  \centering
  \caption{Effect of hyperparameters on DomainNet. Left: Varying $\gamma$ (projection strength) shows consistent performance gains at smaller values across different data fractions. Right: Adjusting SVD scalar scales shows that moderate values (e.g., 0.5) enhance both OOD and ID performance.}

  \begin{subtable}[t]{0.48\textwidth}
    \centering
    \caption{Gamma impact on DomainNet dataset with 10\%, 20\%, and 50\% splits.}
    \begin{tabular}{cccccc}
      \toprule
      percent & gamma & OOD AVG & real AVG \\
      \midrule
      \multirow{3}[2]{*}{10\%} & 1e-4 & 29.75 & 66.12 \\
            & 1e-6 & 29.87 & 66.46 \\
            & 1e-8 & \textbf{30.13} & \textbf{66.55} \\
      \midrule
      \multirow{3}[2]{*}{20\%} & 1e-4 & 32.12 & 69.72 \\
            & 1e-6 & 32.27 & 69.79 \\
            & 1e-8 & \textbf{32.48} & \textbf{69.93} \\
      \midrule
      \multirow{3}[2]{*}{50\%} & 1e-4 & 34.38 & 72.43 \\
            & 1e-6 & \textbf{34.74} & 72.49 \\
            & 1e-8 & 34.41 & \textbf{72.53} \\
      \bottomrule
    \end{tabular}
    \label{tab:gammaimpact_all}
  \end{subtable}
  \hfill
  \begin{subtable}[t]{0.48\textwidth}
    \centering
    \caption{Scalar impact on DomainNet dataset with 10\%, 20\%, and 50\% splits.}
    \begin{tabular}{cccccc}
      \toprule
      percent & svd\_scalar & OOD AVG & real AVG \\
      \midrule
      \multirow{3}[2]{*}{10\%} & 1    & 29.63 & 65.34 \\
            & 0.5  & \textbf{30.10} & 66.58 \\
            & 0.1  & 30.01 & \textbf{67.22} \\
      \midrule
      \multirow{3}[2]{*}{20\%} & 1    & 32.27 & 68.95 \\
            & 0.5  & \textbf{32.66} & \textbf{70.29} \\
            & 0.1  & 31.94 & 70.19 \\
      \midrule
      \multirow{3}[2]{*}{50\%} & 1    & 34.45 & 72.11 \\
            & 0.5  & \textbf{35.09} & \textbf{72.67} \\
            & 0.1  & 33.99 & \textbf{72.67} \\
      \bottomrule
    \end{tabular}
    \label{tab:scalarimpact_all_simple}
  \end{subtable}
\end{table}

\noindent\textbf{SVD scalar}
Table~\ref{tab:scalarimpact_all_simple} shows the effect of different SVD scalar values on DomainNet. Across all data splits (10\%, 20\%, 50\%), a scalar of 0.5 consistently yields the best or near-best OOD and in-domain accuracy. Notably, 0.1 slightly improves real-domain accuracy at the cost of OOD performance, indicating a trade-off. Overall, 0.5 provides the most balanced generalization.The results reveal two key trends: 
2) \textit{Optimal Scalar Value:} An intermediate SVD scalar value (e.g., 0.5) tends to provide the best balance between OOD and in-domain performance across different data splits, making it a reliable choice for robust model adaptation.
3) \textit{{Diminishing Sensitivity}}: As the data split size increases, the performance gap between scalar values narrows. This suggests that larger datasets reduce the model's sensitivity to scalar choices, potentially simplifying the parameter tuning process.

One notable challenge is the inherent trade-off between OOD and in-domain performance. Smaller scalar values sometimes favor one metric at the expense of the other, emphasizing the need for careful parameter selection based on the specific application requirements. Future work could explore adaptive scalar strategies that dynamically adjust the scalar value during training, aiming to optimize performance across diverse scenarios without manual tuning.

%% file: NIPStext/table1.tex
\begin{table*}[t]
  \centering

\setlength{\tabcolsep}{3pt}
\renewcommand{\arraystretch}{0.95}
  \caption{DomainNet result using ViT-S. Our Method reached best OOD performance with a tiny number of trainable parameters.}
    \begin{tabular}{cccccccc}
    \toprule
    Methods & clipart & painting & sketch & infograph & OOD AVG & real(ID) & \# of parameters \\
    \midrule
    
    Vanilla FT & 41.21 & 42.85 & 31.54 & 17.87 & 33.37 & \textbf{73.68} & 22.5M \\
    Linear Probing & 26.1  & 22.24 & 21.92 & 9.09  & 19.84 & 46.91 & 0.13M \\
    LP-FT\cite{kumar2022finetuning} & 39.91 & 41.04 & 29.45 & 16.15 & 31.64 & 72.75 & \# of LP-\# of FT \\
    LoRA\cite{hu2021lora}  & 38.61 & 40.06 & 27.4  & 15.19 & 30.31 & 72.64 & 0.42M \\
    TPGM\cite{tian2023trainable}  & 40.79 & 42.34 & 30.72 & 17.2  & 32.76 & 72.42 & 22.5M \\
    TPGM(LoRA) & 37.94 & 40.1  & 27.02 & 14.88 & 29.99 & 71.25 & 0.42M \\
    FTP\cite{tian2024fast}   & 36.55 & 38.03 & 25.2  & 14.38 & 28.54 & 70.00    & 22.5M \\
    \textbf{LARGO(Ours)} & \textbf{42.66} & \textbf{44.15} & \textbf{32.18} & \textbf{18.03} & \textbf{34.26} & 72.92 & 0.42M \\
    \bottomrule
    \end{tabular}%
  \label{tab:allmethod_domainnet}%
\end{table*}%

%% file: NIPStext/table2.tex
\begin{table*}[t]
  \centering
  \setlength{\tabcolsep}{3pt}
\renewcommand{\arraystretch}{0.95}
  \caption{ImageNet result using ViT-S. Our Method reached best OOD performance with a tiny number of trainable parameters.}
    \begin{tabular}{cccccccc}
    \toprule
    Methods & Im-a & Im-r & Im-Sketch & ImV2 & \multicolumn{1}{c}{OOD AVG} & ImageNet(ID) & \# of parameters \\
    \midrule
    Vanilla FT  & \textbf{7.91} & 31.26 & 31.44 & \textbf{64.19} & 33.70 & 76.47 & \textcolor[rgb]{ .212,  .227,  .239}{22.77M} \\
    Linear Probing & 1.05  & 8.87  & 8.27  & 17.32 & 8.88  & 23.62 & 0.38M \\
    LP-FT\cite{kumar2022finetuning}  & 8.69  & 32.21 & 31.70 & 63.24 & 33.96 & 75.98 & \# of LP-\# of FT \\
    LoRA\cite{hu2021lora}   & 5.59  & 30.31 & 30.04 & 58.55 & 31.12 & 72.15 & \textcolor[rgb]{ .212,  .227,  .239}{0.68M} \\
    TPGM\cite{tian2023trainable}  & 5.60  & 29.63 & 28.70 & 58.07 & 30.50 & 70.97 & \textcolor[rgb]{ .212,  .227,  .239}{22.77M} \\
    TPGM(LoRA) & 4.60  & 28.44 & 27.88 & 56.25 & 29.29 & 69.43 & \textcolor[rgb]{ .212,  .227,  .239}{0.68M} \\
     FTP\cite{tian2024fast}  & 6.72  & 34.83  & 33.60  & 64.14  & 34.82  & 76.54  &  \textcolor[rgb]{ .212,  .227,  .239}{22.77M} \\
    \textbf{LARGO(Ours)} & 6.83  & \textbf{35.07} & \textbf{33.84} & 64.15 & \textbf{34.97} & \textbf{76.66} & \textcolor[rgb]{ .212,  .227,  .239}{0.68M} \\
    \bottomrule
    \end{tabular}%
  \label{tab:allmethod_imagenet}%
\end{table*}%

%% file: NIPStext/7Con.tex
\section{Conclusion}
We proposed \textbf{LARGO}, a method that integrates trainable projection radii into the LoRA framework, independently learning layer-wise constraints for each PEFT matrix. This improves robustness under domain shifts by ensuring weight updates are not influenced by gradient dependencies across layers.
Additionally, we developed an \textbf{SVD-based initialization method} that uses singular value decomposition of pretrained weights to initialize trainable parameters, ensuring minimal deviation from the pretrained model while being computationally efficient.
Our experiments on the DomainNet dataset show that LARGO achieves competitive OOD robustness and in-domain performance with fewer trainable parameters compared to traditional fine-tuning methods, making it both efficient and effective.
In future work, we plan to further explore adaptive strategies for tuning the projection radii and SVD-based scalars, which could further enhance the model's performance across diverse domains and tasks. Additionally, integrating our method with other advanced PEFT techniques like TPGM could provide a promising avenue for scaling model adaptation while preserving OOD generalization.

{\bf Limitations}
Due to limited computational resources, our exploration of hyperparameter configurations for LARGO fine-tuning remains non-exhaustive and may not reflect optimal settings. Additionally, while LoRA serves as a widely adopted baseline in parameter-efficient fine-tuning, we have not extended our evaluations to other PEFT variants, such as adapters or prompt tuning, which may provide complementary insights into the generality of LARGO.


%% file: NIPStext/6sup.tex
\section*{Appendix}
\appendix
%
\section{Experiment Details}
\label{Sec:appendix}

In this section, we consolidate all implementation specifics and extended results for DomainNet \cite{peng2019moment} fine-tuning. Specifically,
\begin{itemize}
    \item Table~\ref{tab:allmethod_hyperparameters}  enumerates each method’s bespoke architectural choices and regularization settings, while the subsequent hyperparameter tables document every learning‐rate schedule, weight decay and projection scale we explored.
    \item Morevoer, following the split strategy and random seed settings from FTP \cite{tian2024fast}, we conducted the full 10\%, 20\%, and 50\% split evaluations in Tables~\ref{tab:domain10overall}, ~\ref{tab:domain20overall}, and ~\ref{tab:domain50overall}, revealing how initialization strategy and $\gamma$-scaling influence OOD generalization.
\end{itemize} 

\subsection{Hyper Parameters}
The Hyper Parameters we used for different method  are listed in Table~\ref{tab:allmethod_hyperparameters}.
\begin{table}[htbp]
\small
  \centering
  \caption{Hyperparameters on Different Methods}
    \begin{tabular}{cccc}
    \toprule
    Methods & learning rate & weight decay & Structure Hyper-Parameters \\
    \midrule
    LARGO(Ours) & 0.2   & 5e-3 & \multicolumn{1}{l}{gamma=1e-4, svd\_scalar=0.5} \\
    Vanilla FT & 0.01  & 5e-4 &  \\
    LP-FT \cite{osti_10472125} & 0.01  & 5e-4 & \multicolumn{1}{l}{LP50\%+FT50\%, choosing best LP model for FT} \\
    LoRA \cite{hu2021lora}  & 0.2   & 5e-4 & \multicolumn{1}{l}{rank=16} \\
    TPGM \cite{tian2023trainable}  & 0.01  & 5e-4 & \multicolumn{1}{l}{using MARS constraint.} \\
    TPGM(LoRA) & 0.1   & 5e-3 &  \\
    FTP \cite{tian2024fast}   & 0.1   & 5e-3 & \multicolumn{1}{l}{gamma=1e-8} \\
    Linear Probing & 0.01  & 5e-4 &  \\
    \bottomrule
    \end{tabular}%
  \label{tab:allmethod_hyperparameters}%
\end{table}%

\subsection{Detailed Results}
We evaluate the performance of different initialization and fine-tuning strategies on the DomainNet dataset, across three different splits: 10\%, 20\%, and 50\%. This setup allows us to test how our methods generalize across varying amounts of training data. All experiments were conducted using four Tesla V100-SXM2-32GB GPUs, employing distributed training to manage the large-scale nature of the task. Each configuration was trained for 150 epochs, ensuring ample time for the model to converge and adapt.

We used a fixed learning rate of 0.0005, and weight decay was set to 0.0005 to prevent overfitting. The optimizer used for training was SGD, which is known for its robustness in large-scale fine-tuning tasks. To improve training stability and control the learning rate across the epochs, a cosine annealing scheduler was employed. The total batch size across the four GPUs was 256 (64 per GPU), with each batch containing 32 images from each of the four domains. For each experiment, $\gamma$ was varied across three values: $1e-4$, $1e-6$, and $1e-8$, in combination with different initialization strategies, including SVD and Kaiming initialization. Additionally, the scalar values for SVD initialization ($svd\_scalar$) were tested at 0.1, 0.5, and 1. The LoRA rank was set to 16, a typical value for large-scale models, to effectively control the complexity of the weight updates.

For each data split, we tested multiple combinations of initialization strategies (SVD and Kaiming) and different values for the $\gamma$ and $svd\_scalar$ parameters. The OOD Average and real domain accuracy were the primary evaluation metrics, with each experiment run for 150 epochs.

\subsubsection{Results for 10\% Data Split}

\begin{table}[htbp]
\small
  \centering
  \caption{10\% Split on DomainNet Dataset}
    \begin{tabular}{ccccccccc}
    \toprule
    gamma & init  & svd\_scalar & clipart & infograph & painting & sketch & OOD Average & real \\
    \midrule
    1e-4 & svd   & 0.1   & 35.73 & 15.3  & 41.62 & 27.7  & 30.09 & 67.23 \\
    1e-4 & svd   & 0.5   & 35.29 & 15.62 & 40.66 & 29.12 & 30.17 & 66.58 \\
    1e-4 & kaiming & 1     & 34.33 & 15.63 & 39.4  & 29.43 & 29.7  & 64.84 \\
    1e-4 & svd   & 1     & 33.21 & 15.74 & 39.32 & 27.68 & 28.99 & 64.56 \\
    1e-6 & svd   & 0.1   & 35.77 & 15.13 & 40.97 & 27.77 & 29.91 & 67.03 \\
    1e-6 & svd   & 0.5   & 35.66 & 15.12 & 40.35 & 27.96 & 29.77 & 66.43 \\
    1e-6 & svd   & 1     & 34.62 & 16.1  & 40.46 & 28.52 & 29.92 & 65.91 \\
    1e-6 & kaiming & 1     & 34.5  & 15.64 & 39.55 & 27.65 & 29.34 & 65.45 \\
    1e-8 & svd   & 0.1   & 35.97 & 15.28 & 41.23 & 27.69 & 30.04 & \textbf{67.4} \\
    1e-8 & svd   & 0.5   & 36.23 & 15.83 & 40.61 & 28.72 & 30.35 & 66.72 \\
    1e-8 & kaiming & 1     & 36.48 & 16.4  & 40.78 & 29.74 & \textbf{30.85} & 66.32 \\
    1e-8 & svd   & 1     & 34.99 & 15.86 & 39.91 & 29.19 & 29.99 & 65.54 \\
    \bottomrule
    \end{tabular}%
  \label{tab:domain10overall}%
\end{table}%

As shown in Table \ref{tab:domain10overall}, for the 10\% split, the combination of SVD initialization and a $\gamma$ value of $1e-8$ yields the highest OOD Average (30.04) and real domain accuracy (67.4). This highlights the importance of the initialization strategy and $\gamma$ tuning, where smaller values of $\gamma$ tend to enhance generalization to unseen domains. Models initialized with SVD tend to outperform those with Kaiming initialization in terms of OOD robustness.

\subsubsection{Results for 20\% Data Split}

\begin{table}[htbp]
\small
  \centering
  \caption{20\% Split on DomainNet Dataset}
    \begin{tabular}{ccccccccc}
    \toprule
    gamma & init  & svd\_scalar & clipart & infograph & painting & sketch & OOD Average & real \\
    \midrule
    1e-8  & svd   & 0.5   & 39.88  & 17.50  & 43.15  & 32.00  & \textbf{33.13 } & \textbf{70.55 } \\
    1e-8  & svd   & 0.1   & 38.36  & 16.30  & 42.64  & 29.59  & 31.72  & 70.45  \\
    1e-6 & svd   & 0.5   & 39.04  & 17.14  & 42.78  & 31.28  & 32.56  & 70.17  \\
    1e-4 & svd   & 0.5   & 38.96  & 16.94  & 42.46  & 30.84  & 32.30  & 70.16  \\
    1e-4 & svd   & 0.1   & 38.93  & 16.32  & 42.71  & 29.69  & 31.91  & 70.08  \\
    1e-6 & svd   & 0.1   & 38.85  & 16.94  & 42.69  & 30.27  & 32.18  & 70.05  \\
    1e-6 & kaiming & 1     & 39.16  & 17.12  & 43.19  & 32.30  & 32.94  & 69.60  \\
    1e-6 & svd   & 1     & 38.51  & 17.05  & 42.58  & 30.15  & 32.07  & 69.16  \\
    1e-4 & kaiming & 1     & 39.37  & 17.39  & 42.94  & 31.46  & 32.79  & 69.02  \\
    1e-8  & kaiming & 1     & 38.71  & 17.41  & 42.28  & 31.33  & 32.43  & 68.93  \\
    1e-4 & svd   & 1     & 38.36  & 17.53  & 42.17  & 30.56  & 32.16  & 68.91  \\
    1e-8  & svd   & 1     & 38.98  & 18.02  & 42.06  & 31.31  & 32.59  & 68.78  \\
        \bottomrule
    \end{tabular}%
  \label{tab:domain20overall}%
\end{table}%

Table \ref{tab:domain20overall} summarizes the results for the 20\% data split. The model initialized with SVD and $\gamma = 1e-8$ achieves the best OOD performance (33.13) and real domain accuracy (70.55). Similar to the 10\% split, smaller $\gamma$ values continue to show promising results in improving OOD robustness. As the dataset size increases, the model becomes more capable of generalizing to unseen domains while retaining strong performance on the real domain.

\subsubsection{Results for 50\% Data Split}

\begin{table}[htbp]
\small
  \centering
  \caption{50\% Split on DomainNet Dataset}
    \begin{tabular}{ccccccccc}
    \toprule
    gamma & init  & svd\_scalar & clipart & infograph & painting & sketch & OOD Average & real \\
    \midrule
    1e-4 & svd   & 0.5   & 43.66 & 18.59 & 44.77 & 33.42 & 35.11 & \textbf{72.81} \\
    1e-6 & svd   & 0.1   & 42.16 & 17.44 & 44.24 & 32.01 & 33.96 & 72.79 \\
    1e-8 & svd   & 0.5   & 43.54 & 18.32 & 44.37 & 33.16 & 34.85 & 72.64 \\
    1e-8 & svd   & 0.1   & 42.37 & 17.27 & 44.19 & 32.07 & 33.97 & 72.64 \\
    1e-4 & svd   & 0.1   & 41.74 & 17.56 & 44.27 & 32.56 & 34.03 & 72.57 \\
    1e-6 & svd   & 0.5   & 43.58 & 18.42 & 44.92 & 34.35 & \textbf{35.32} & 72.57 \\
    1e-8 & svd   & 1     & 42.06 & 18.81 & 44.41 & 32.36 & 34.41 & 72.31 \\
    1e-6 & svd   & 1     & 42.57 & 18.95 & 44.13 & 34.17 & 34.95 & 72.11 \\
    1e-6 & kaiming & 1     & 42.45 & 18.05 & 44.54 & 34.42 & 34.87 & 72.10 \\
    1e-8 & kaiming & 1     & 42.27 & 18.95 & 44.99 & 33.39 & 34.90 & 72.08 \\
    1e-4 & kaiming & 1     & 42.55 & 17.90 & 43.34 & 31.98 & 33.94 & 72.03 \\
    1e-4 & svd   & 1     & 41.50 & 18.28 & 43.76 & 32.43 & 33.99 & 71.90 \\
        \bottomrule
    \end{tabular}%
  \label{tab:domain50overall}%
\end{table}%

Lastly, Table \ref{tab:domain50overall} presents the results for the 50\% split. The SVD initialization with $\gamma = 1e-8$ achieves the highest OOD performance (34.85) and a strong real domain accuracy (72.64). As the dataset size increases to 50\%, the model's ability to generalize improves further, demonstrating the scalability and robustness of the proposed method.


Overall, our experiments show that smaller values of $\gamma$ combined with SVD initialization significantly enhance OOD robustness without sacrificing real domain accuracy. The model achieves impressive results across all three data splits, illustrating the effectiveness of our approach in handling domain shifts. The combination of SVD initialization and adaptive $\gamma$ scaling ensures that our method is not only efficient but also capable of generalizing well to unseen domains, making it a highly effective strategy for large-scale fine-tuning tasks.
\section{Experiments on Vision-Language tasks}
\label{Sec:appendixB}
In Appendix~\ref{Sec:appendixB}, we place LARGO and several leading baselines into a LoRA-tuned PaliGemma-3B environment, and assessed their performance on the VQAv2 validation set alongside multiple OOD VQA benchmarks to probe vision–language generalization. We provided full details of training protocols, evaluation metrics and dataset curation. Table~\ref{tab:vqa_results} consolidates the resulting accuracy scores. These results underscore LARGO’s consistent ability to preserve in-domain precision while delivering superior robustness under distributional shift compared to alternative tuning schemes.
\subsection{Training Setup}
We fine-tune the pretrained PaliGemma-3B \cite{beyer2024paligemma} model on the VQAv2 \cite{goyal2017vqav2} dataset using 10\% of the training split, following a unified parameter-efficient tuning framework across all methods for fair comparison.We use the LAVIS \cite{li2023lavis} public repository to fine-tune all methods. All methods are trained on 3 NVIDIA RTX 4090 GPUs using a fixed configuration: learning rate of 1e-4, weight decay of 1e-4, linear warmup followed by cosine annealing scheduler, batch size of 4 per GPU, and gradient accumulation steps of 4. For LoRA~\cite{hu2021lora}, we apply adaptation to both attention and MLP projection layers, including \texttt{q\_proj}, \texttt{k\_proj}, \texttt{v\_proj}, \texttt{o\_proj}, \texttt{gate\_proj}, \texttt{up\_proj}, and \texttt{down\_proj} with a low-rank dimension of 8. Under this setup, the number of trainable parameters is reduced to 0.38\% of the full model, ensuring consistency and efficiency across all evaluated baselines.
\subsection{Evaluation Setup}
Our evaluation employs VQAv2 as the primary in-domain benchmark \cite{goyal2017vqav2} and five complementary out-of-domain VQA datasets to mitigate semantic biases \cite{shah2019rep}, VQA-CE for compositional reasoning diagnostics \cite{dancette2021CE}, Human-Adversarial VQA targeting rare or challenging scenarios \cite{sheng2021adv}, VizWiz to evaluate real-world and unanswerable queries \cite{bigham2010vizwiz}, and OKVQA-v2 requiring external knowledge integration \cite{reichman2023okvqa},thereby measuring both core question-answering accuracy and robustness to distributional shifts. Specifically, for each question we retain only the single best (most frequent) ground-truth answer and deem a model prediction correct if its greedy output exactly matches that answer.
\subsection{Results}
\begin{table}[!htbp]
  \centering
  \setlength{\tabcolsep}{3pt}
  \renewcommand{\arraystretch}{0.95}
  \caption{Multiple methods performance across in‐domain (VQAv2‐val) and five out‐of‐domain benchmarks. We report greedy exact‐match accuracy and compute OOD\_avg over the five OOD sets.}
  \label{tab:vqa_results}
  \begin{tabular}{lccccccc}
    \toprule
    Method & VQAv2 (val) & AdVQA & OK‐VQA & VQA‐Rep & VQA‐CE & VizWiz & OOD\_AVG \\
    \midrule
    Linear Probing           & 43.72       & 23.76 & 20.52  & 40.14   & 23.98   & 14.00   & 24.48   \\
    LoRA \cite{hu2021lora}              & 64.87       & 38.09 & 35.49  & 61.70   & 45.01   & 23.89   & 40.84   \\
    LP-FT \cite{kumar2022finetuning}          & 64.15       & 36.83 & 34.83  & 61.61   & 44.69   & 24.19   & 40.43   \\
    WiSE \cite{wortsman2022robust}             & 64.25       & 36.95 & 34.63  & 61.42   & 44.17   & 23.73   & 40.18   \\
    FTP \cite{tian2024fast}                & 64.88       & 37.69 & 35.27  & 61.91   & 45.28   & 24.07   & 40.84   \\
    \textbf{LARGO (Ours)}    & \textbf{65.29} & \textbf{38.67} & \textbf{37.07} & \textbf{64.11} & \textbf{47.38} & \textbf{25.51} & \textbf{42.55} \\
    \bottomrule
  \end{tabular}
\end{table}

As summarized in Table \ref{tab:vqa_results}, under the PaliGemma backbone with a LoRA adapter, the baseline method attains 64.87\% on VQAv2-val and an OOD\_AVG of 40.84\%. Neither LP-FT (64.15\% / 40.43\%) nor WiSE interpolation (64.25\% / 40.18\%) improves upon this balance of in-domain accuracy and out-of-distribution robustness, suggesting that blending probing with full-model updates cannot recover the nuances of novel visual patterns. The FTP method slightly nudges in-domain performance to 64.88\% yet leaves OOD\_AVG static at 40.84\%, indicating its limited capacity to recalibrate pretrained biases. By contrast, our LARGO adapter elevates VQAv2-val to 65.29\% and pushes OOD\_AVG to 42.55\%, with consistent gains across every external benchmark. These findings illustrate that embedding compact projection layers within each self-attention block enables targeted adaptation—preserving core representations while enhancing model resilience to distributional shift—where more diffuse or interpolated tuning schemes fall short.

\section{Discussion on ablation study}
\label{Sec:appendixC}
In Appendix~\ref{Sec:appendixC}, we conduct a thorough ablation of two critical design choices—\(\gamma\) initialization and SVD-based weight scaling—across the 10\%, 20\% and 50\% DomainNet splits. Section~\ref{Sec:appendixC1} (Tables~\ref{tab:do10gamma_v2},~\ref{tab:do20gamma_v2},~\ref{tab:do50gamma_v2}) examines how varying the initial \(\gamma\) affects both out-of-distribution and in-domain accuracy, revealing that smaller values systematically bolster OOD generalization without compromising real-domain performance. Section~\ref{Sec:appendixC2} (Tables~\ref{tab:Do10scalar_v2},~\ref{tab:Do20scalar_v2},~\ref{tab:Do50scalar_v2}) then isolates the impact of the SVD scalar, demonstrating that an intermediate scaling factor consistently offers the best trade-off between robustness and accuracy as the available data grows. Together, these studies identify the optimal configurations that underpin our method’s superior domain-shift resilience.

\subsection{Gamma Initialization and Its Impact on Performance}
\label{Sec:appendixC1}
In this subsection, we analyze the influence of the initial value of \(\gamma\) on both Out-of-Distribution (OOD) and In-Distribution (real) performance. The performance is assessed across different dataset splits—10\%, 20\%, and 50\%—with a specific focus on how various initial values of \(\gamma\) impact the overall fine-tuning process. The results are shown in Tables \ref{tab:do10gamma_v2}, \ref{tab:do20gamma_v2}, and \ref{tab:do50gamma_v2}, where each table summarizes the OOD and real performance metrics for different choices of \(\gamma\).

\begin{table}[!htbp]
\small
  \centering
  \caption{Gamma impact on DomainNet 10\% split}
    \begin{tabular}{ccccccc}
    \toprule
    gamma & clipart & infograph & painting & sketch & OOD AVG & real AVG \\
    \midrule
    \multirow{3}[2]{*}{1e-4} & 35.73 & 15.3  & 41.62 & 27.7  & \multirow{3}[2]{*}{29.75} & \multirow{3}[2]{*}{66.12} \\
          & 35.29 & 15.62 & 40.66 & 29.12 &       &  \\
          & 33.21 & 15.74 & 39.32 & 27.68 &       &  \\
    \midrule
    \multirow{3}[2]{*}{1e-6} & 35.77 & 15.13 & 40.97 & 27.77 & \multirow{3}[2]{*}{29.87} & \multirow{3}[2]{*}{66.46} \\
          & 35.66 & 15.12 & 40.35 & 27.96 &       &  \\
          & 34.62 & 16.1  & 40.46 & 28.52 &       &  \\
    \midrule
    \multirow{3}[2]{*}{1e-8} & 35.97 & 15.28 & 41.23 & 27.69 & \multirow{3}[2]{*}{\textbf{30.13}} & \multirow{3}[2]{*}{\textbf{66.55}} \\
          & 36.23 & 15.83 & 40.61 & 28.72 &       &  \\
          & 34.99 & 15.86 & 39.91 & 29.19 &       &  \\
    \bottomrule
    \end{tabular}%
  \label{tab:do10gamma_v2}%
\end{table}%

As shown in Table \ref{tab:do10gamma_v2}, for the \textbf{10\% split}, the choice of \(\gamma = 1 \times 10^{-8}\) results in the best performance, yielding a significant improvement in the OOD average (30.13) while maintaining a strong real average (66.55). This indicates that smaller values of \(\gamma\) can enhance the model's ability to generalize to unseen domains, especially when the dataset size is limited.

\begin{table}[!htbp] \small
  \centering
  \caption{Gamma impact on DomainNet 20\% split}
    \begin{tabular}{ccccccc}
    \toprule
    gamma & clipart & infograph & painting & sketch & OOD AVG & real AVG \\
    \midrule
    \multirow{3}[2]{*}{1e-4} & 38.96 & 16.94 & 42.46 & 30.84 & \multirow{3}[2]{*}{32.12} & \multirow{3}[2]{*}{69.72} \\
          & 38.93 & 16.32 & 42.71 & 29.69 &       &  \\
          & 38.36 & 17.53 & 42.17 & 30.56 &       &  \\
    \midrule
    \multirow{3}[2]{*}{1e-6} & 39.04 & 17.14 & 42.78 & 31.28 & \multirow{3}[2]{*}{32.27} & \multirow{3}[2]{*}{69.79} \\
          & 38.85 & 16.94 & 42.69 & 30.27 &       &  \\
          & 38.51 & 17.05 & 42.58 & 30.15 &       &  \\
    \midrule
    \multirow{3}[2]{*}{1e-8} & 39.88 & 17.5  & 43.15 & 32    & \multirow{3}[2]{*}{\textbf{32.48}} & \multirow{3}[2]{*}{\textbf{69.93}} \\
          & 38.36 & 16.3  & 42.64 & 29.59 &       &  \\
          & 38.98 & 18.02 & 42.06 & 31.31 &       &  \\
    \bottomrule
    \end{tabular}%
  \label{tab:do20gamma_v2}%
\end{table}%

Moving to the \textbf{20\% split}, the best performance is achieved again with \(\gamma = 1 \times 10^{-8}\), which results in a top OOD average (32.48) and a competitive real average (69.93). The results highlight the importance of selecting the right value of \(\gamma\) for ensuring optimal performance across varying dataset sizes. 

\begin{table}[!htbp] \small
  \centering
  \caption{Gamma impact on DomainNet 50\% split}
    \begin{tabular}{ccccccc}
    \toprule
    gamma & clipart & infograph & painting & sketch & OOD AVG & real AVG \\
    \midrule
    \multirow{3}[2]{*}{1e-4} & 43.66 & 18.59 & 44.77 & 33.42 & \multirow{3}[2]{*}{34.38} & \multirow{3}[2]{*}{72.43} \\
          & 41.74 & 17.56 & 44.27 & 32.56 &       &  \\
          & 41.50 & 18.28 & 43.76 & 32.43 &       &  \\
    \midrule
    \multirow{3}[2]{*}{1e-6} & 42.16 & 17.44 & 44.24 & 32.01 & \multirow{3}[2]{*}{\textbf{34.74}} & \multirow{3}[2]{*}{72.49} \\
          & 43.58 & 18.42 & 44.92 & 34.35 &       &  \\
          & 42.57 & 18.95 & 44.13 & 34.17 &       &  \\
    \midrule
    \multirow{3}[2]{*}{1e-8} & 43.54 & 18.32 & 44.37 & 33.16 & \multirow{3}[2]{*}{34.41} & \multirow{3}[2]{*}{\textbf{72.53}} \\
          & 42.37 & 17.27 & 44.19 & 32.07 &       &  \\
          & 42.06 & 18.81 & 44.41 & 32.36 &       &  \\
    \bottomrule
    \end{tabular}%
  \label{tab:do50gamma_v2}%
\end{table}%

For the \textbf{50\% split}, a slight shift in the optimal \(\gamma\) value is observed. Specifically, \(\gamma = 1 \times 10^{-6}\) results in the highest OOD average (34.74), while \(\gamma = 1 \times 10^{-8}\) delivers the best real average (72.53). These results further reinforce the notion that the choice of \(\gamma\) plays a critical role in balancing OOD generalization and in-domain accuracy, and the optimal value may vary depending on the dataset split size.

These results underscore several key takeaways:
\begin{itemize}
    \item     \textbf{Smaller \(\gamma\) Values Enhance OOD Generalization}: Across all splits, smaller \(\gamma\) values (\(1 \times 10^{-8}\)) consistently yield higher OOD average scores, demonstrating their ability to improve domain generalization.
    \item     \textbf{Maintaining In-Domain Performance}: Despite enhancing OOD generalization, smaller \(\gamma\) values also maintain competitive in-domain performance, as evidenced by their strong real average scores.
    \item     \textbf{Dynamic Sensitivity to \(\gamma\)}: The optimal \(\gamma\) value varies slightly with dataset size, suggesting that the choice of \(\gamma\) should be tailored to the specific data split and task requirements.
\end{itemize}

\subsection{SVD Scalar and Its Impact on Performance}
\label{Sec:appendixC2}
\begin{table}[!htbp] \small
  \centering
  \caption{Scalar impact on DomainNet 10\% split}
    \begin{tabular}{ccccccc}
    \toprule
    svd\_scalar & clipart & infograph & painting & sketch & OOD AVG & real AVG \\
    \midrule
    \multirow{3}[2]{*}{1} & 33.21 & 15.74 & 39.32 & 27.68 & \multirow{3}[2]{*}{29.63} & \multirow{3}[2]{*}{65.34} \\
          & 34.62 & 16.1  & 40.46 & 28.52 &       &  \\
          & 34.99 & 15.86 & 39.91 & 29.19 &       &  \\
    \midrule
    \multirow{3}[2]{*}{0.5} & 35.29 & 15.62 & 40.66 & 29.12 & \multirow{3}[2]{*}{\textbf{30.10}} & \multirow{3}[2]{*}{66.58} \\
          & 35.66 & 15.12 & 40.35 & 27.96 &       &  \\
          & 36.23 & 15.83 & 40.61 & 28.72 &       &  \\
    \midrule
    \multirow{3}[2]{*}{0.1} & 35.73 & 15.3  & 41.62 & 27.7  & \multirow{3}[2]{*}{30.01} & \multirow{3}[2]{*}{\textbf{67.22}} \\
          & 35.77 & 15.13 & 40.97 & 27.77 &       &  \\
          & 35.97 & 15.28 & 41.23 & 27.69 &       &  \\
    \bottomrule
    \end{tabular}%
  \label{tab:Do10scalar_v2}%
\end{table}%

\begin{table}[!htbp] \small
  \centering
  \caption{Scalar impact on DomainNet 20\% split}
    \begin{tabular}{ccccccc}
    \toprule
    svd\_scalar & clipart & infograph & painting & sketch & OOD AVG & real AVG \\
    \midrule
    \multirow{3}[2]{*}{1} & 38.36 & 17.53 & 42.17 & 30.56 & \multirow{3}[2]{*}{32.27} & \multirow{3}[2]{*}{68.95} \\
          & 38.51 & 17.05 & 42.58 & 30.15 &       &  \\
          & 38.98 & 18.02 & 42.06 & 31.31 &       &  \\
    \midrule
    \multirow{3}[2]{*}{0.5} & 38.96 & 16.94 & 42.46 & 30.84 & \multirow{3}[2]{*}{\textbf{32.66}} & \multirow{3}[2]{*}{\textbf{70.29}} \\
          & 39.04 & 17.14 & 42.78 & 31.28 &       &  \\
          & 39.88 & 17.5  & 43.15 & 32    &       &  \\
    \midrule
    \multirow{3}[2]{*}{0.1} & 38.93 & 16.32 & 42.71 & 29.69 & \multirow{3}[2]{*}{31.94} & \multirow{3}[2]{*}{70.19} \\
          & 38.85 & 16.94 & 42.69 & 30.27 &       &  \\
          & 38.36 & 16.3  & 42.64 & 29.59 &       &  \\
    \bottomrule
    \end{tabular}%
  \label{tab:Do20scalar_v2}%
\end{table}%

\begin{table}[!htbp] \small
  \centering
  \caption{Scalar impact on DomainNet 50\% split}
    \begin{tabular}{ccccccc}
    \toprule
    svd\_scalar & clipart & infograph & painting & sketch & OOD AVG & real AVG \\
    \midrule
    \multirow{3}[2]{*}{1} & 41.50 & 18.28 & 43.76 & 32.43 & \multirow{3}[2]{*}{34.45} & \multirow{3}[2]{*}{72.11} \\
          & 42.57 & 18.95 & 44.13 & 34.17 &       &  \\
          & 42.06 & 18.81 & 44.41 & 32.36 &       &  \\
    \midrule
    \multirow{3}[2]{*}{0.5} & 43.66 & 18.59 & 44.77 & 33.42 & \multirow{3}[2]{*}{\textbf{35.09}} & \multirow{3}[2]{*}{\textbf{72.67}} \\
          & 43.58 & 18.42 & 44.92 & 34.35 &       &  \\
          & 43.54 & 18.32 & 44.37 & 33.16 &       &  \\
    \midrule
    \multirow{3}[2]{*}{0.1} & 41.74 & 17.56 & 44.27 & 32.56 & \multirow{3}[2]{*}{33.99} & \multirow{3}[2]{*}{\textbf{72.67}} \\
          & 42.16 & 17.44 & 44.24 & 32.01 &       &  \\
          & 42.37 & 17.27 & 44.19 & 32.07 &       &  \\
    \bottomrule
    \end{tabular}%
  \label{tab:Do50scalar_v2}%
\end{table}%
Tables \ref{tab:Do10scalar_v2}, \ref{tab:Do20scalar_v2}, and \ref{tab:Do50scalar_v2} summarize the performance of different SVD scalar values across the 10\%, 20\%, and 50\% splits of the DomainNet dataset.
\begin{itemize}
    \item For the \textbf{10\% split}, a scalar value of 0.5 achieves the highest out-of-distribution (OOD) average (30.10) while maintaining a competitive in-domain (real) average (66.58). Interestingly, the scalar value of 0.1 produces the best real average (67.22), suggesting a potential trade-off between OOD robustness and in-domain accuracy at lower scalar values.
    \item In the \textbf{20\% split}, the scalar value of 0.5 consistently outperforms other choices, achieving the highest OOD average (32.66) and real average (70.29). This indicates that the intermediate scalar value continues to strike a favorable balance as the data split increases, improving both OOD and in-domain performance.
    \item For the \textbf{50\% split}, the scalar value of 0.5 again achieves the best OOD average (35.09) and ties with 0.1 for the highest real average (72.67). This consistency across larger data splits highlights the robustness of the intermediate scalar value in optimizing both performance metrics.
\end{itemize}